%% file: main.tex
\documentclass[letterpaper]{article} % DO NOT CHANGE THIS
\usepackage[]{aaai23}  % DO NOT CHANGE THIS
\usepackage{times}  % DO NOT CHANGE THIS
\usepackage{helvet}  % DO NOT CHANGE THIS
\usepackage{courier}  % DO NOT CHANGE THIS
\usepackage[hyphens]{url}  % DO NOT CHANGE THIS
\usepackage{graphicx} % DO NOT CHANGE THIS
\usepackage{natbib}  % DO NOT CHANGE THIS AND DO NOT ADD ANY OPTIONS TO IT
\usepackage{caption} % DO NOT CHANGE THIS AND DO NOT ADD ANY OPTIONS TO IT
\usepackage{algorithm}
\usepackage{algorithmic}

\usepackage{newfloat}
\usepackage{listings}
\DeclareCaptionStyle{ruled}{labelfont=normalfont,labelsep=colon,strut=off} % DO NOT CHANGE THIS
\floatstyle{ruled}
\newfloat{listing}{tb}{lst}{}
\floatname{listing}{Listing}
\input{macros}

\title{Conflicting Interactions Among Protection Mechanisms for Machine Learning Models}
\author {
    Sebastian Szyller,\textsuperscript{\rm 1}
    N. Asokan \textsuperscript{\rm 2 \rm 1}
}
\affiliations {
    \textsuperscript{\rm 1} Aalto University\\
    \textsuperscript{\rm 2} University of Waterloo\\
    contact@sebszyller.com, asokan@acm.org
}
\begin{document}

\maketitle

\input{0abstract}
\input{1introduction}
\input{2background}
\input{3problem}
\input{4approach}
\input{5exp-setup}
\input{6evaluation}
\input{7addressing}
\input{8discussion}
\input{9related}
\input{10conclusion}
\input{11acknowledgements}

\bibliography{bibliography}

\input{12appendix}

\end{document}

%% file: macros.tex
\usepackage{amsmath,amsfonts}
\usepackage{dsfont}
\usepackage{graphicx}
\usepackage{textcomp}
\usepackage{xcolor}
\usepackage{colortbl}
\usepackage{xspace}
\usepackage{tabularx}
\usepackage{cleveref}
\usepackage{multirow}
\usepackage{subcaption}
\usepackage[inline]{enumitem}
\usepackage{soul}
\usepackage{xurl}

\newcolumntype{Y}{>{\centering\arraybackslash}X}

\newcommand{\adv}{\ensuremath{\mathcal{A}}\xspace}
\newcommand{\victim}{\ensuremath{\mathcal{V}}\xspace}

\newcommand{\fv}{\ensuremath{F_{\mathcal{V}}}\xspace}
\newcommand{\fa}{\ensuremath{F_{\mathcal{A}}}\xspace}

\newcommand{\dataset}{\ensuremath{\mathcal{D}}\xspace}
\newcommand{\traindata}{\ensuremath{\mathcal{D}_\text{TR}}\xspace}
\newcommand{\testdata}{\ensuremath{\mathcal{D}_\text{TE}}\xspace}
\newcommand{\triggerset}{\ensuremath{\mathcal{D}_\text{T}}\xspace}
\newcommand{\raddata}{\ensuremath{\mathcal{D}_\text{RAD}}\xspace}

\newcommand{\metric}{\ensuremath{\phi}\xspace}
\newcommand{\metricacc}{\ensuremath{\phi_\text{ACC}}\xspace}
\newcommand{\metricwm}{\ensuremath{\phi_\text{WM}}\xspace}
\newcommand{\metricadv}{\ensuremath{\phi_\text{ADV}}\xspace}
\newcommand{\metricrad}{\ensuremath{\phi_\text{RAD}}\xspace}
\newcommand{\metricdi}{\ensuremath{\phi_\text{DI}}\xspace}

\newcommand{\advt}{\textsf{ADVTR}\xspace}
\newcommand{\di}{\textsf{DI}\xspace}
\newcommand{\radioactive}{\textsf{RADDATA}\xspace}
\newcommand{\diffpr}{\textsf{DP}\xspace}

\newcommand{\dpsgd}{\textsf{DPSGD}\xspace}
\newcommand{\wm}{\textsf{WM}\xspace}
\newcommand{\ours}{\textsf{OURS}\xspace}
\newcommand{\cecg}{\cellcolor{gray!30}\xspace}
\newcommand{\cecp}{\cellcolor{purple!30}\xspace}
\newcommand{\cecb}{\cellcolor{blue!30}\xspace}
\newcommand{\cecgr}{\cellcolor{green!20}\xspace}

\newcommand{\hypo}{\ensuremath{\mathcal{H}_0}\xspace}
\newcommand{\hypoalt}{\ensuremath{\mathcal{H}^*_0}\xspace}

\newcommand{\good}[1]{\textcolor{teal}{\underline{#1}}}
\newcommand{\bad}[1]{\textcolor{red}{#1}}
\newcommand{\techreport}[0]{}

\newcommand{\argmax}{\operatornamewithlimits{argmax}}

%% file: 0abstract.tex
%auto-ignore
\begin{abstract}
Nowadays, systems based on machine learning (ML) are widely used in different domains.
Given their popularity, ML models have become targets for various attacks.
As a result, research at the intersection of security/privacy and ML has flourished. Typically such work has focused on individual types of security/privacy concerns and mitigations thereof.

However, in real-life deployments, an ML model will need to be \emph{protected against several concerns simultaneously}.
A \emph{protection mechanism} optimal for a specific security or privacy concern may interact negatively with mechanisms intended to address other concerns. Despite its practical relevance, the potential for such conflicts has not been studied adequately.

In this work, we first provide a framework for analyzing such \emph{conflicting interactions}.
We then focus on systematically analyzing pairwise interactions between protection mechanisms for one concern, \emph{model and data ownership verification}, with two other classes of ML protection mechanisms: \emph{differentially private training}, and \emph{robustness against model evasion}.
We find that several pairwise interactions result in conflicts.

We also explore potential approaches for avoiding such conflicts. First, we study the effect of hyperparameter relaxations, finding that there is no sweet spot balancing the performance of both protection mechanisms.
Second, we explore whether modifying one type of protection mechanism (ownership verification) so as to decouple it from factors that may be impacted by a conflicting mechanism (differentially private training or robustness to model evasion) can avoid conflict.
We show that this approach can indeed avoid the conflict between ownership verification mechanisms when combined with differentially private training, but has no effect on robustness to model evasion. We conclude by identifying the gaps in the landscape of studying interactions between other types of ML protection mechanisms.
\end{abstract}

%% file: 1introduction.tex
% auto-ignore
\section{Introduction}
\label{sec:introduction}

Machine learning (ML) models constitute valuable intellectual property.
They are also increasingly deployed in risk-sensitive domains. As a result, various security and privacy requirements for ML model deployment have become apparent.
This, in turn, has led to substantial recent research at the intersection of machine learning and security/privacy.
%
%There are different types of threats against ML models.
The research community largely focuses on individual types of security/privacy threats and ways to defend against them.
This facilitates iterative improvements, and allows practitioners to evaluate the benefit of any new approaches.

In this work, we argue that in realistic deployment setting, multiple security/privacy concerns need to be considered \emph{simultaneously}.
Therefore, any \emph{protection mechanism} for a particular concern, needs to be tested together with defences against \emph{other common concerns}.
We show that when deployed together, ML protection mechanisms may not work as intended due to \emph{conflicting interactions} among them.

We claim the following contributions:
\begin{enumerate}[label=\textbf{\arabic*)}]
    \item \label{contribution-1} We highlight the importance of understanding \emph{conflicting interactions} among ML protection mechanisms, and provide a framework for studying it (\Cref{sec:problem-statement}).
    \item \label{contribution-2} We use our framework to analyse the interaction between \emph{model ownership verification mechanisms} with two other types of protection mechanisms: \emph{differentially private training} and \emph{adversarial training}. We provide a theoretical justification (\Cref{sec:approach}) for each potential pairwise conflict, and evaluate it empirically (\Cref{sec:exp-setup,sec:evaluation}).
    \item \label{contribution-3} We explore whether conflicts can be avoided by changing (a) the hyperparameters of each protection mechanism, or (b) the design of the mechanism itself (\Cref{sec:addressing}).
\end{enumerate}

%% file: 2background.tex
%auto-ignore
\section{Background}
\label{sec:background}

We begin with  a brief background on relevant topics.

\subsection{Machine Learning}

The goal of a ML classification model \fv trained on some dataset \traindata is to perform well on the given classification task according to some metric \metric measured on a test set \testdata.
The whole dataset is denoted as $\dataset = \{ \traindata, \testdata \}$.
An individual record consists of an input $x$ and the corresponding label $y$.
Throughout this work, we use the accuracy metric $\metricacc(\fv, \testdata)$ to assess a model \fv using \testdata:
\begin{equation}
    \metricacc(\fv, \testdata) = \frac{1}{|\testdata|} \sum\limits_{x \in \testdata} \mathds{1} (\hat{\fv}(x) = y).
\end{equation}

where $\fv(x)$ is the full probability vector and $\hat{\fv}(x)$ is the most likely class.

\subsection{Ownership Verification}
In a \emph{white-box model stealing} attack an adversary \adv obtains an identical copy of \fv belonging to a victim \victim, e.g., by breaking into a device, or bribing an employee.
Watermarking can be used to deter white-box model stealing~\cite{zhang2018protecting,adi2018turning,uchida2017embedding,rouhani2019deepsigns}.

On the other hand, in \emph{black-box model extraction} attacks~\cite{papernot2017practical,juuti2019prada,orekondy2018knockoff,tramer2016stealing,correia2018copycat,jagielski2020highfidelity,carlini2020cryptanalytic,pal2019publicdata,takemura2020rnns}, \adv ``steals'' \fv by sending queries, recording responses, and using them to train a \emph{surrogate} model \fa.

Model extraction defences try to either \emph{detect}~\cite{juuti2019prada,atli2020boogeyman,quiring2018forgottensib} or \emph{slow down}~\cite{orekondy20prediction,dziedzic2022increasing,lee2018defending} the attack but cannot \emph{prevent} it.
Adversarial watermarking~\cite{szyller2021dawn} can \emph{deter} extraction attacks by forcing a watermark into \fa, or by ensuring that a watermark transfers from \fv to \fa~\cite{jia2021entangled}.

\noindent\textbf{Backdoor watermarking}~\cite{zhang2016understanding,adi2018turning,szyller2021dawn} (\wm) allows \victim to embed an out-of-distribution watermark (``trigger set'') \triggerset during training.
\triggerset is chosen so that it does not interfere with the primary learning task, and is difficult to discover.
Effectiveness of watermarks can be assessed via the accuracy of \fv on \triggerset:
\begin{equation}
    \metricwm(\fv, \triggerset) = \frac{1}{|\triggerset|} \sum\limits_{x \in \triggerset} \mathds{1} (\hat{\fv}(x) = y).
\end{equation}

For \fv with $m$ classes, the verification confidence is~\cite{szyller2021dawn}:
\begin{equation}\label{eq:wmconf}
    V =
    \sum_{i=0}^{\lfloor e \times \vert \triggerset \vert \rfloor}
    \binom{\vert \triggerset \vert}{i}
    \times
    \left( \dfrac{m-1}{m} \right) ^{i}
    \times
    \left( \dfrac{1}{m} \right) ^{\vert \triggerset \vert-i}
\end{equation}
where $e = 1 - \metricwm(\fv, \triggerset)$ is the tolerated error rate for $V$.
If \victim suspects a model \fa to be a copy of \fv, it can query \fa with \triggerset to verify ownership.

\noindent\textbf{Radioactive data}~\cite{sablayrolles2020radioactive} (\radioactive) is a \emph{dataset watermarking} technique allowing \victim to identify models trained using their datasets.
\radioactive embeds imperceptible perturbations in a subset of the training images $x_\phi = x + \phi$ which constitute a watermark \raddata.
The perturbations are crafted iteratively using an optimization procedure similar to adversarial example search.
Its goal is to align the images with a particular part of the manifold.
Intuitively, \raddata is more difficult to train on than the (clean) counterpart, and subsequently, more difficult to classify correctly with high confidence.
Hence, any model trained on \raddata would perform better on it.

The effectiveness of \radioactive can be measured using a white-box approach based on hypothesis testing or a black-box based on the loss difference.
In this work, we use the black-box approach as it performs better in the original work~\cite{tekgul2022datawm}.
For a loss function $\mathcal{L}(\fv(x), y)$ the black-box ownership verification metric is defined as:
\begin{align}
    \metricrad(\fv, \raddata)
    &= \frac{1}{|\raddata|} \times \nonumber \\
    \times \sum\limits_{\{x, x_\phi\} \in \raddata} &\mathds{1}
    ( \mathcal{L}(\fv(x), y) -  \mathcal{L}(\fv(x_\phi), y))
\end{align}

$\metricrad > 0$ indicates that that \fv was trained on \raddata.
The higher the value, the more confident the verification.

\noindent\textbf{Dataset inference}~\cite{maini2021datasetinference} (\di) is a model \emph{fingerprinting} technique.
It assumes that \fv was trained on a private training dataset.
It exploits the fact that an \fa extracted from \fv would learn similar features as \fv.

To create the fingerprint, \victim extracts the feature embeddings from \fv that characterise their prediction margin using several distance metrics.
The embeddings for \traindata and \testdata are used to train a distinguisher.
During verification, \victim queries \fa with a subset of \traindata, $D_{ver} \subset \traindata$.
\fa is deemed stolen if the distances are similar to \fv with sufficient confidence, under a hypothesis test.
The verification is successful if the p-value (\metricdi) is below a certain threshold.

The embeddings can be obtained using a white-box technique (\emph{MinGD}) or a black-box one (\emph{Blind Walk}).
In this work, we use Blind Walk approach as it performs better in the original work.
The success of \di (\metricdi) is measured by comparing the embeddings using a hypothesis test.
Distinguishable embeddings indicate that the model is stolen.

\subsection{Model Evasion}

In a model evasion attack~\cite{biggio2013evasion,szegedy2013intriguingproperties}, an adversary \adv aims to craft an \emph{adversarial example} $x_\gamma = x + \gamma$ such that it is misclassified by \fv, $\hat{\fv}(x) \ne \hat{\fv}(x_\gamma)$.
Typically, the attack is restricted to produce inputs that are within $\gamma$ distance (according  to some distance measure, typically $L_{2}$ or $L_\infty$) from the originals.
\techreport{
\footnote{In perception related tasks, like image classification, it is commonly assumed that that the perturbation is imperceptible. However, in practice, the adversary will inject $\gamma$ that causes maximum harm.}}

\emph{Adversarial training} (\advt) is %a modification to the standard training procedure and is
designed to provide robustness against adversarial examples.
During training, each clean sample $x$ is replaced with an adversarial example $x_\gamma$.
Robustness can be measured by calculating the accuracy of the model on the adversarial test set:
\begin{equation}
    \metricadv(\fv, \testdata) = \frac{1}{|\testdata|} \sum\limits_{x \in \testdata} \mathds{1} (\hat{\fv}(x + \gamma) = y).
\end{equation}

\advt is successful if \metricadv is high (ideally the same as \metricacc), and \metricacc does not deteriorate.

There exist many techniques for crafting adversarial examples that in turn can be used in \advt.
We use \emph{projected gradient descent}~\cite{madry2017towards} (PGD) a popular optimization technique for crafting adversarial examples.

\subsection{Differential Privacy}

Differential privacy~\cite{dwork2006dp} (\diffpr) bounds \adv's capability to infer information about any individual record in \traindata.
The learning algorithm $A$ satisfies $(\epsilon, \delta)$-differential privacy if for any two datasets $D$, $D'$ that differ in one record, and any set of models $Q$:
\begin{equation}
    Pr[A(D) \in Q] \le e^\epsilon Pr[A(D') \in Q] + \delta
\end{equation}

$\epsilon$ corresponds to the privacy budget, and $\delta$ is the probability mass for events where the privacy loss is larger than $e^\epsilon$.
Together, these two can be considered as $\phi_\text{DP}$;

In this work,  we use the most popular algorithm for differentially private training \dpsgd~\cite{abadi2016dpsgd}.
%

%% file: 3problem.tex
%auto-ignore

\section{Problem Statement}
\label{sec:problem-statement}

The efficacy of a ML protection mechanism $M$ with hyperparameters $\theta_M$ is measured by an associated metric $\metric_M$. Many ML protection mechanisms tend to decrease \metricacc.
Therefore, the goal of a mechanism is to maximise \emph{both} $\metric_M$  and \metricacc.
An individual instantiation $m$ of $M$ applied to the model $F$, $m(F, \theta_M)$, seeks to maximise both $\metric_M$ and $\metricacc$:
\begin{equation}
    m^* = \argmax_m \{ \metric_M (m(F, \theta_M), \cdot),  \metricacc (m(F, \theta_M), \cdot) \}
\end{equation}

Consequently, the instantiation is effective iff.:
\begin{enumerate}
    \item the difference, $\Delta_{\metric_M} = |\metric_M(F) - \metric_M(m(F,\theta_M))|$,
          for a given metric is \emph{above} a threshold $t_{\metric_M}$: $\Delta_{\metric_M} > t_{\metric_M}$.
    \item the difference, $\Delta_{\metricacc} = |\metricacc(F) - \metricacc(m(F, \theta_M))|$,
          is \emph{below} a threshold $t_{\metricacc}$: $\Delta_{\metricacc} < t_{\metricacc}$.
\end{enumerate}
For instance, \victim may find that \advt with $\Delta_{\metricadv} < 0.3$ and $\Delta_{\metricacc} > 0.2$ is unacceptable.
In reality, acceptable thresholds are application- and deployment-specific.

\techreport{
The instantiation applies either to the model or to the training procedure.
We omit this in the notation, as well as the hyperparameters associated with the model for the sake of brevity.
}

We can extend this to a combination of mechanisms $C = \{ M_1, M_2, \dots, M_n \}$.
A combination $c$ of multiple individual instantiations $\{ m_1, m_2, \dots, m_n \}$ applied to
$F$, $c(F, \theta)$
where $\theta = \{ \theta_{M_1}, \theta_{M_2}, \dots, \theta_{M_n} \}$ is effective iff. all metrics
${\metric}_{M_1, M_2, \dots, M_n}$ and $\metricacc$ are sufficiently high:
\begin{align}
    c^* =
        \argmax_{c=\{ m_1, m_2, \dots, m_n \}} \{
        &\metric_{M_1} (c(F, \theta), \cdot),  \nonumber\\
        &\metric_{M_2} (c(F, \theta), \cdot),  \nonumber\\
        &\dots \nonumber\\
        &\metric_{M_n} (c(F, \theta), \cdot),  \nonumber\\
        &\metricacc (c(F, \theta), \cdot) \}
\end{align}

In other words, the combination is effective iff. \emph{all} $\Delta_{\metric_M} \in \{ \Delta_{\metric_{M_1}}, \Delta_{\metric_{M_2}}, \dots, \Delta_{\metric_{M_n}} \}$ are \emph{below} their corresponding thresholds, and $\Delta_{\metricacc} < t_{\metricacc}$.
Unlike for a single mechanism, here, $\Delta_M$ is calculated with $c$ applied: $\Delta_{\metric} = |\metric(m(F, \theta_M), \cdot) - \metric(c(F, \theta), \cdot)|$.

Given a pair of mechanisms $C_{M_1, M_2} = \{ M_1, M_2\}$, our goal is to determine if there exists an effective combination of instantiations $c_{m_1, m_2}$ such that $\Delta_{\metric_{M_1}} < t_{\metric_{M_1}}$, $\Delta_{\metric_{M_2}} < t_{\metric_{M_2}}$ and $\Delta_{\metricacc} < t_{\metricacc}$.
Subsequently, a pair is in \emph{conflict} if any of these three inequalities does not hold.
For a single mechanism, its threshold denotes required gain; while for a combination it corresponds to a maximum decrease.

Crucially, for a given pair $C_{M_1, M_2}$, both $\metric_{M}${s} have an upper bound: $\metric_{M}(c(F, \theta), \cdot) \le \metric_{M}(m(F, \theta_M), \cdot)$; and similarly, \metricacc has an upper bound:
\begin{align}\label{eq:minacc}
    \metricacc(c(F, \theta)) \le \min(&\metricacc(m_1(F, \theta_{M_1}), \cdot), \nonumber\\
                                      &\metricacc(m_2(F, \theta_{M_2}), \cdot))
\end{align}

\noindent\textbf{Choosing thresholds.}
%we established that a mechanism $M_1$ is ineffective if %$\Delta_{\metric_{M_1}} > t_{\metric_{M_1}}$.
%It begs the question what the appropriate thresholds are.
In this work, we use the following thresholds to decide if a combination of mechanisms is ineffective:
%we consider $M_1$ ineffective if after combining it with $M_2$,
$\Delta_{\metricacc} > 10pp$, or:
\begin{enumerate*}[label=\arabic*)]
\item \wm: $\Delta_{\metricwm} > 30pp$;
\item \advt: $\Delta_{\metricadv} > 10pp$;
\item \di: $\metricdi > 10^{-3}$;
\item \radioactive: $\metricrad < 10^{-2}$;
\end{enumerate*}

Note that \dpsgd has a tight bound for the given $(\epsilon, \delta)$~\cite{milad2021advinst}.
However, we consider increasing $\epsilon \times 1.5$ too permissive for the purpose of the changes discussed in~\Cref{sec:addressing}.

%% file: 4approach.tex
%auto-ignore
\section{Conflicting Interactions}
\label{sec:approach}

We consider \emph{pair-wise} interactions between protection mechanisms that allow for ownership verification \{\wm, \di, \radioactive\} and techniques based on strong regularisers \{\advt, \dpsgd\}.
We first explain why a given pair may conflict. In~\Cref{sec:evaluation} we verify our hypotheses empirically.

\subsection{Pair-wise Conflicts}
\noindent\textbf{\dpsgd with \wm.}
\wm relies on overfitting \fv to the trigger set \triggerset (memorisation) while simultaneously trying to learn the primary dataset \traindata.
In turn, the gradient norm of \triggerset is high which is necessary to provide sufficient signal.
On the other hand, \dpsgd relies on two primary mechanisms that limit the contribution of individual samples: 1) clipping and 2) adding noise to the gradients.
These induce strong regularization on \traindata.
Hence, these two techniques impose contradictory objectives for the training to optimize for.
Therefore, we conjecture that they conflict.

\noindent\textbf{\advt with \wm.}
\advt provides a strong regularising property to the model that typically results in a decrease in $\metricacc$ for non-trivial values of $\gamma$.
Similarly to the interaction with \dpsgd, we suspect that the regularization induced by \advt will harm the embedding of \triggerset.
Furthermore, \triggerset introduces \emph{pockets} of OOD data, and it was shown that \triggerset is indistinguishable from \traindata in the final layer~\cite{szyller2021dawn}.
These may make finding adversarial examples easier.

\noindent\textbf{\dpsgd or \advt with \di.}
\di relies on the fact that \fa derived from \fv has similar decision boundaries.
Because \dpsgd limits the contribution of individual samples at the gradient level, the decision boundaries of \fv trained with and without it may differ.
\advt changes the decision boundary around each training record, and may conflict with \di.

\noindent\textbf{\advt with \radioactive.}
\radioactive relies on the optimization procedure that is similar to finding adversarial examples.
Hence, we expect \advt to prevent \raddata from being embedded.
It is unclear if the presence of \raddata is going to negatively impact \metricadv.

\noindent\textbf{\dpsgd with \radioactive.}
Like \wm, \radioactive requires embedding information that differs from \traindata.
We expect \dpsgd to limit the contribution of \raddata and decrease, or invalidate \metricrad.

\subsection{Result Significance}
For all experiments we measure the statistical significance of the result.
We test \metricacc, and then each $\metric_{M}$ separately.
We first conduct a \emph{t-test} under the null hypothesis \hypo of \emph{equivalent population distributions} with $\alpha=0.05$.
Next, if \hypo is rejected, we conduct a \emph{two one-sided test} to see  if the result falls within the \emph{equivalence bounds} (for which we use the abovementioned thresholds).
Here, the null hypothesis \hypoalt is reversed and we assume \emph{non-equivalence}, $\alpha^*=0.05$.
\hypoalt is rejected if the results fall within the bound.
For both, we use Welch's t-test since sample variances are unequal.

%% file: 5exp-setup.tex
%auto-ignore
\section{Experimental Setup}
\label{sec:exp-setup}

\noindent\textbf{Datasets.}
we use three benchmark datasets for our evaluation.
MNIST~\cite{lecun2010mnist} contains $60,000$ train and $10,000$ test grayscale images of digits.
The corresponding label is the digit presented in the image.
FashionMNIST~\cite{xiao2017fashionmnist} (FMNIST) contains $60,000$ train and $10,000$ test grayscale images of articles of clothing, divided into $10$ classes.
The corresponding label is the piece of clothing presented in the image.
CIFAR10~\cite{krizhevsky2009cifar10} contains $50,000$ train and $10,000$ test RGB images which depict miscellaneous animals or vehicles, divided into $10$ classes.

For MNIST and FMNIST, we use each as the out-of-distribution dataset from which watermarks for the other are drawn.
For CIFAR10, we follow prior work~\cite{szyller2021dawn} in using GTSRB~\cite{stallkamp2011german} as the source of out-of-distribution watermarks. GTSRB is a traffic sign dataset that contain $39,209$ train and $12,630$ test RGB images, divided into 43 classes.
%, we use it as an %out-of-distribution watermark dataset for CIFAR10.
%We do not use it as a benchmark dataset because it has similar %complexity to CIFAR10, and hence does not provide any additional %information.
%\newtext{}{We use MNIST and FMNIST to draw watermarks for each other.}

\noindent\textbf{Models and training.}
for MNIST and FMNIST, we use a simple 4-layer CNN.
We train the models for $25$ epochs for the baselines, and $100$ for experiments with protection mechanisms deployed.
For all experiments we use the initial learning rate of $0.001$ and maximum learning rate of $0.005$ with a one-cycle scheduler~\cite{smith2018onecycle}.

For CIFAR10, we use a ResNet20.
We train the models for $100$ epochs for the baselines, and $200$ for experiments with protection mechanisms deployed.
Similarly to MNIST and FMNIST case, we use a one-cycle scheduler.
However, we use the initial learning rate of $0.005$ and maximum starting learning of $0.1$.
We summarise these details in~\Cref{tab:setup-details}.

\begin{table}
    \centering
    \caption{Datasets, model architecture, and training. Learning rate reported as: \textit{starting}/\textit{maximum}; number of epochs reported as: \textit{baseline (unprotected)}/\textit{protected}.}
    \label{tab:setup-details}
    % \resizebox{\linewidth}!{
    \scriptsize
    \begin{tabularx}{\linewidth}{|Y|Y|Y|c|c|c|}\hline
        Dataset   & $|\traindata|$ & $|\testdata|$ & Architecture & Learning rate           & Epochs  \\ \hline
        MNIST     & $50,000$       & $10,000$      & 4-layer CNN  & $0.001/0.005$ & 25/100  \\
        FMNIST    & $60,000$       & $10,000$      & 4-layer CNN  & $0.001/0.005$ & 25/100  \\
        CIFAR10   & $50,000$       & $10,000$      & ResNet20     & $0.005/0.1$   & 100/200 \\
        GTSRB     & $39,209$       & $12,630$      & -            & -                       & -       \\ \hline
    \end{tabularx}
    % }
\end{table}

\techreport{
In total, we trained over $1,000$ models to guarantee the consistency of the results.}
For baselines models, and those with only one mechanism, training was repeated five times; for pair-wise comparisons, training was repeated ten times.
All training was done, on a workstation with two NVIDIA RTX 3090, Threadripper 3960X, and 128 GB of RAM.
We used the PyTorch library to train the models.
We use official repositories of techniques that we evaluate:
\wm\footnote{\url{https://github.com/ssg-research/dawn-dynamic-adversarial-watermarking-of-neural-networks}},
\di\footnote{\url{https://github.com/cleverhans-lab/dataset-inference}},
\radioactive\footnote{\url{https://github.com/facebookresearch/radioactive_data}}.
The code for this project is on GitHub\footnote{\url{https://github.com/ssg-research/conflicts-in-ml-protection-mechanisms}}.

%% file: 6evaluation.tex
%auto-ignore
\section{Evaluation}
\label{sec:evaluation}

\setlength{\extrarowheight}{.2em}
\begin{table*}[!ht]
    \centering
    \caption{Baseline models without any protection mechanisms (No Def.), and with a single mechanism deployed.
    We provide \metricacc and the corresponding metric for each mechanism.
    Results are averaged over 5 runs; we report the mean and standard deviation rounded to two decimal places (three for \radioactive).}
    \label{tab:baselines}
    \scriptsize
    \begin{tabularx}\linewidth{|Y|Y|Y|Y|Y|Y|Y|Y|c|c|}\hline
            \multirow{2}{*}{Dataset} & No Def.       & \multicolumn{2}{c|}{\advt}    & \dpsgd              & \multicolumn{2}{c|}{\wm}      & \multicolumn{2}{c|}{\radioactive} & \multicolumn{1}{c|}{\di} \\
                                     & \metricacc    & \metricacc & \metricadv       & \metricacc          & \metricacc    & \metricwm     & \metricacc    & \metricrad        & \metricdi \\\hline
            MNIST                    & 0.99$\pm$0.00 & 0.99$\pm$0.00 & 0.95$\pm$0.00 & 0.98$\pm$0.00       & 0.99$\pm$0.00 & 0.97$\pm$0.01 & 0.98$\pm$0.00 & 0.284$\pm$0.001   & $<10^{-30}$ \\
            FMNIST                   & 0.91$\pm$0.00 & 0.87$\pm$0.00 & 0.69$\pm$0.00 & 0.86$\pm$0.01       & 0.87$\pm$0.02 & 0.99$\pm$0.02 & 0.88$\pm$0.01 & 0.191$\pm$0.002   & $<10^{-30}$ \\
            CIFAR10                  & 0.92$\pm$0.00 & 0.88$\pm$0.00 & 0.82$\pm$0.00 & 0.38$\pm$0.00       & 0.82$\pm$0.00 & 0.97$\pm$0.02 & 0.85$\pm$0.00 & 0.202$\pm$0.001   & $<10^{-30}$ \\\hline
    \end{tabularx}
\end{table*}

\setlength{\extrarowheight}{.2em}
\begin{table}[!h]
    \centering
    \caption{Summary of the hyperparameters for each protection mechanism.}
    \label{tab:hyperparameters}
    \scriptsize
    \begin{tabularx}{\columnwidth}{|Y|c|c|c|c|c|c|}\hline
        \multirow{2}{*}{Dataset} & \advt           & \multicolumn{3}{c|}{\dpsgd}  & \wm             & \radioactive \\
                                 & $\gamma$        & $\epsilon$ & $\delta$  & $c$ & $|\triggerset|$ & $\raddata$ \% \\\hline
        MNIST                    & 0.25            & 3          & $10^{-6}$ & 1.0 & 100             & 10\%\\
        FMNIST                   & 0.25            & 3          & $10^{-6}$ & 1.0 & 100             & 10\%\\
        CIFAR10                  & 10/255          & 3          & $10^{-5}$ & 1.0 & 100             & 10\%\\\hline
    \end{tabularx}
\end{table}

We report on our experiments for studying how ownership verification mechanisms interact with \diffpr (\Cref{sec:evaluation-impact-of-dp}) and \advt (\Cref{sec:evaluation-impact-of-adv-training}).
We color-code all results to convey potential conflicts (e.g., $\bad{0.3}$) or their absence (e.g., $\good{0.7}$).

We used the following hyperparameters: 1) for \dpsgd, clipping norm $c=1.0$, $\epsilon=3$, $\delta=10^{-6}$ for MNIST and FMNIST, and $\delta=10^{-5}$ for CIFAR10; 2) for \advt, $\gamma=0.25$ for MNIST and FMNIST, and $\gamma=10/255$ for CIFAR10; 3) for \wm, $|\triggerset|=100$; for \radioactive, watermarked ratio of $10\%$.
\Cref{tab:baselines} gives the baseline results for all three datasets; \Cref{tab:hyperparameters} summarizes the hyperparameters of each protection mechanism.

Note that we chose the CIFAR10 architecture (ResNet20) that is capable of supporting \wm, \radioactive and \di.
As a result, \metricacc with \dpsgd is relatively low.
There exist mechanisms that achieve higher \metricacc; however, they have additional implications.
We discuss this in~\Cref{sec:discussion-relaxations-schemes}.

\subsection{Impact of Differential Privacy}
\label{sec:evaluation-impact-of-dp}

\Cref{tab:defenses-with-dp} gives the results for combining \dpsgd with \wm, \radioactive, and \di.
In all cases, \metricacc remains close to the single-mechanism baselines.
For MNIST and CIFAR10, \hypo cannot be rejected (not enough evidence to indicate that accuracy differs between the baseline(s) and the jointly protected instances).
For FMNIST \hypo is rejected, but \hypoalt is rejected as well, leading us to conclude that the accuracy is the same within the equivalence bounds.

\noindent\textbf{\wm.}
For all datasets, \metricwm drops significantly.
\hypo is rejected, and \hypoalt is close to $1.0$ for the equivalence bound of $30pp$.
Samples in \triggerset are outliers, and require memorization for successful embedding.
\dpsgd, by design, bounds the contribution of individual samples during training.

\noindent\textbf{\radioactive.}
\dpsgd lowers \metricrad but it remains high enough for successful verification.
Both \hypo and \hypoalt are rejected indicating that while the results are different they are within equivalence bounds.
The regularizing effect of \dpsgd is insufficient to prevent \fv from learning \raddata.
Unlike in \wm, \radioactive modifies the samples such that they align with a few selected carriers.
Hence, multiple samples in \raddata nudge the model in the same direction allowing it to learn the watermark.

\noindent\textbf{\di.}
\di retains high verification confidence (\hypo cannot be rejected).

In summary, \metricacc remains high in all cases.
\wm performance is destroyed, \radioactive has reduced effectiveness but not enough to declare a conflict, and \di has no conflict.

\setlength{\extrarowheight}{.3em}
\begin{table*}[!ht]
    \centering
    \caption{Simultaneous deployment of \dpsgd with \wm, \radioactive and \di.
    \wm drops over $30pp$.
    The loss difference for \radioactive is reduced but still allows for confident verification.
    \di is unaffected.
    \metricacc remains close to the baseline value in all cases.
    Results are averaged over 10 runs; we report the mean and standard deviation rounded to two decimal places (three for \radioactive).
    Green underlined indicates no conflict, red indicates conflict - outside the equivalence bound.}
    \label{tab:defenses-with-dp}
    % \scriptsize
    % \begin{tabularx}\linewidth{|Y|Y|Y|Y|Y|Y|c|c|c|c|c|Y|Y|Y|}\hline
    %         \multirow{3}{*}{Dataset} & \dpsgd        & \multicolumn{4}{c|}{\wm}                                                    & \multicolumn{4}{c|}{\radioactive}                                              & \multicolumn{2}{c|}{\di} \\
    %                                  & Baseline      & \multicolumn{2}{c|}{Baseline} & \multicolumn{2}{c|}{+\dpsgd}                & \multicolumn{2}{c|}{Baseline}   & \multicolumn{2}{c|}{+\dpsgd}                 & Baseline   & +\dpsgd \\
    %                                  & \metricacc    & \metricacc & \metricwm        & \metricacc           & \metricwm            & \metricacc    & \metricrad      & \metricacc           & \metricrad            & \metricdi    & \metricdi     \\\hline
    %         MNIST                    & 0.98$\pm$0.00 & 0.99$\pm$0.00 & 0.97$\pm$0.01 & \good{0.97$\pm$0.00} & \bad{0.36$\pm$0.06}  & 0.98$\pm$0.00 & 0.284$\pm$0.001 & \good{0.97$\pm$0.00} & \good{0.091$\pm$0.01} & $<10^{-30}$ & \good{$<10^{-30}$}  \\
    %         FMNIST                   & 0.86$\pm$0.01 & 0.87$\pm$0.02 & 0.99$\pm$0.02 & \good{0.86$\pm$0.00} & \bad{0.30$\pm$0.05}  & 0.88$\pm$0.01 & 0.191$\pm$0.002 & \good{0.84$\pm$0.01} & \good{0.11$\pm$0.01}  & $<10^{-30}$ & \good{$<10^{-30}$}  \\
    %         CIFAR10                  & 0.38$\pm$0.00 & 0.82$\pm$0.00 & 0.97$\pm$0.02 & \good{0.38$\pm$0.01}  & \bad{0.12$\pm$0.01} & 0.85$\pm$0.00 & 0.202$\pm$0.001 & \good{0.35$\pm$0.01} & \good{0.19$\pm$0.01}  & $<10^{-30}$ & \good{$<10^{-30}$}  \\\hline
    %     \end{tabularx}
    \scriptsize
    \begin{tabularx}\linewidth{|Y|Y|Y|Y|Y|Y|c|c|c|c|c|Y|Y|}\hline
            \multirow{3}{*}{Dataset} & \dpsgd        & \multicolumn{4}{c|}{\wm}                                                    & \multicolumn{4}{c|}{\radioactive}                                              & \di \\
                                     & Baseline      & \multicolumn{2}{c|}{Baseline} & \multicolumn{2}{c|}{+\dpsgd}                & \multicolumn{2}{c|}{Baseline}   & \multicolumn{2}{c|}{+\dpsgd}                 & +\dpsgd \\
                                     & \metricacc    & \metricacc & \metricwm        & \metricacc           & \metricwm            & \metricacc    & \metricrad      & \metricacc           & \metricrad            & \metricdi     \\\hline
            MNIST                    & 0.98$\pm$0.00 & 0.99$\pm$0.00 & 0.97$\pm$0.01 & \good{0.97$\pm$0.00} & \bad{0.36$\pm$0.06}  & 0.98$\pm$0.00 & 0.284$\pm$0.001 & \good{0.97$\pm$0.00} & \good{0.091$\pm$0.01} & \good{$<10^{-30}$}  \\
            FMNIST                   & 0.86$\pm$0.01 & 0.87$\pm$0.02 & 0.99$\pm$0.02 & \good{0.86$\pm$0.00} & \bad{0.30$\pm$0.05}  & 0.88$\pm$0.01 & 0.191$\pm$0.002 & \good{0.84$\pm$0.01} & \good{0.11$\pm$0.01}  & \good{$<10^{-30}$}  \\
            CIFAR10                  & 0.38$\pm$0.00 & 0.82$\pm$0.00 & 0.97$\pm$0.02 & \good{0.38$\pm$0.01}  & \bad{0.12$\pm$0.01} & 0.85$\pm$0.00 & 0.202$\pm$0.001 & \good{0.35$\pm$0.01} & \good{0.19$\pm$0.01}  & \good{$<10^{-30}$}  \\\hline
        \end{tabularx}
\end{table*}

\subsection{Impact of Adversarial Training}
\label{sec:evaluation-impact-of-adv-training}

\Cref{tab:defenses-with-adv} gives the results for combining \advt with \wm, \radioactive, and \di.
In all cases, \metricacc on average remains close to the single-mechanism baselines.
For the equivalence bound of $10pp$, \hypo can and \hypoalt cannot be rejected only for MNIST.
For FMNIST and CIFAR10, \hypoalt gives a p-value of $0.15$ and $0.07$ respectively.
However, a slightly larger threshold $t_{\metricacc}=11pp$, we obtain a p-value of $0.005$ and $0.002$.
Therefore, we deem that the combination does not affect \metricacc enough to declare a conflict based on \metricacc.

\noindent\textbf{\wm.}
\metricwm remains close to the baselines.
However, \metricadv drops at least $10pp$ for FMNIST and CIFAR10; \hypoalt is close to $1.0$.
This is a surprising result because \triggerset{s} are chosen to be far from the distribution of \traindata.
We conjecture that \triggerset is in fact quite close to \traindata in the weight manifold, and because it has random labels, it is easier for the evasion attack to find a perturbation that leads to a misclassification.

\noindent\textbf{\radioactive.}
On the other hand, \radioactive is rendered ineffective while \metricadv stays high.
\metricrad drops close to zero which leads to a low confidence verification.
\hypo is rejected but \hypoalt is not (the result is significantly below $10^{-2}$).
\radioactive relies on replacing some samples in \traindata with samples similar to adversarial examples.
It then exploits the difference in the loss on clean and perturbed samples for dataset ownership verification.
\advt replaces all data in \traindata with an adversarial variant, and hence, invalidates the mechanism used by \radioactive.

\noindent\textbf{\di.}
Similarly to the pairing with \dpsgd, confidence of \di remains high.

In summary, \metricacc remains high in all cases.
\wm and \di remain effective, while \radioactive performance is destroyed.
\metricadv stays high both for \di and \radioactive but is decreased for \wm.
Only \di has no conflict with \advt.
However, we observed that \di can result in false positives when \fv and \fa where trained using \traindata from the same distribution, even though, \fa is benign.
We discuss this further in~\Cref{sec:limitations-and-fps} and~\Cref{app:fps}.

\setlength{\extrarowheight}{.3em}
\begin{table*}[!ht]
    \centering
    \caption{Simultaneous deployment of \advt with \wm, \radioactive and \di.
    \advt does not interfere with \wm, \metricwm remains high;
    however, \metricadv drops at least $10pp$ for FMNIST and CIFAR10.
    \radioactive is rendered ineffective as \metricadv drops almost to zero.
    \di is unaffected.
    \metricacc remains close to the baseline value in all cases.
    Results are averaged over 10 runs; we report the mean and standard deviation rounded to two decimal places (three for \radioactive).
    Green underlined indicates no conflict, red indicates conflict - outside the equivalence bound.}
    \label{tab:defenses-with-adv}
    \scriptsize
    \begin{tabularx}\linewidth{|Y|Y|Y|Y|Y|Y|Y|c|c|c|c|c|Y|Y|}\hline
        \multirow{3}{*}{Dataset} & \multicolumn{5}{c|}{\wm}                                                                             & \multicolumn{5}{c|}{\radioactive}                                                                     & \di \\
                                 & \multicolumn{2}{c|}{Baseline} & \multicolumn{3}{c|}{+\advt}                                          & \multicolumn{2}{c|}{Baseline}   & \multicolumn{3}{c|}{+\advt}                                         & +\advt  \\
                                 & \metricacc       & \metricwm  & \metricacc            & \metricwm             & \metricadv           & \metricacc & \metricrad         & \metricacc           & \metricrad            & \metricadv           & \metricdi     \\\hline
        MNIST                    & 0.99$\pm$0.00 & 0.97$\pm$0.01 & \good{0.97$\pm$0.02}  & \good{0.99$\pm$0.01}  & \good{0.88$\pm$0.09} & 0.98$\pm$0.00 & 0.284$\pm$0.001 & \good{0.94$\pm0.01$} & \bad{0.001$\pm$0.001} & \good{0.95$\pm$0.01} & \good{$<10^{-30}$}  \\
        FMNIST                   & 0.87$\pm$0.02 & 0.99$\pm$0.02 & \good{0.80$\pm$0.06}  & \good{0.99$\pm$0.00}  & \bad{0.51$\pm$0.11}  & 0.88$\pm$0.01 & 0.191$\pm$0.002 & \good{0.87$\pm0.02$} & \bad{0.000$\pm$0.001} & \good{0.69$\pm$0.02} & \good{$<10^{-30}$} \\
        CIFAR10                  & 0.82$\pm$0.00 & 0.97$\pm$0.02 & \good{0.78$\pm$0.00}  & \good{0.97$\pm$0.01}  & \bad{0.65$\pm$0.01}  & 0.85$\pm$0.00 & 0.202$\pm$0.001 & \good{0.81$\pm0.01$} & \bad{0.003$\pm$0.002} & \good{0.81$\pm$0.01} & \good{$<10^{-30}$}  \\\hline
    \end{tabularx}
\end{table*}

%% file: 7addressing.tex
\section{Addressing the Conflicts}
\label{sec:addressing}

Having established that there are multiple instances of conflicting interactions among ML protection mechanisms,
we now explore how we might avoid conflicts.

First, we investigated whether settling for a weaker protection guarantee of one mechanism meaningfully boosts the performance of the other.
However, the changes either do not sufficiently improve any metric, or require significantly lowering the protection guarantee.
See~\Cref{app:hyperparams} for details.

Second, we separated the training objective of \wm and from the regularization imposed by \advt and \dpsgd, and check if it allows the model to recover some of its original effectiveness.
So far, we used the mechanisms without differentiating between \traindata, and \triggerset or \raddata.
Instead, one could apply these mechanisms only to the primary training task with \traindata, and use a separate parameter optimizer for \triggerset or \radioactive.
We evaluate such modification for the pairs in conflict: 1) \wm with \advt; 2) \radioactive with \advt; 3) \wm with \dpsgd.

\noindent\textbf{\wm with \advt}.
We observed a minor improvement in \metricacc.
However, it does not substantially improve \metricadv, although, it does reduce the standard deviation across runs.
\metricwm remains high as in the previous experiments (\Cref{tab:relaxed-advt}).

\setlength{\extrarowheight}{.2em}
\begin{table*}[ht]
    \centering
    \caption{Training on \traindata with \advt and without on \triggerset.
    The change does not result in any meaningful improvement. Green underlined indicates no conflict; red indicates conflict.}
    \label{tab:relaxed-advt}
    % \resizebox{2.\columnwidth}!{
    \scriptsize
    \begin{tabularx}\linewidth{|Y|Y|Y|Y|Y|Y|Y|Y|Y|Y|Y|Y|Y|}\hline
            \multirow{3}{*}{Dataset} & \multicolumn{2}{c|}{\advt}    & \multicolumn{8}{c|}{\wm}                                                                            \\
                                     & \multicolumn{2}{c|}{Baseline} & \multicolumn{2}{c|}{Baseline} & \multicolumn{3}{c|}{+\advt}                                     & \multicolumn{3}{c|}{+\advt Relaxed}\\
                                     & \metricacc & \metricadv       & \metricacc       & \metricwm  & \metricacc            & \metricwm             & \metricadv          & \metricacc           & \metricwm             & \metricadv\\\hline
            MNIST                    & 0.99$\pm$0.00 & 0.95$\pm$0.00 & 0.99$\pm$0.00 & 0.97$\pm$0.01 & \good{0.97$\pm$0.02}  & \good{0.99$\pm$0.01}  & \good{0.88$\pm$0.09} & \good{0.97$\pm$0.01} & \good{0.99$\pm$0.01}  & \good{0.89$\pm$0.01}\\
            FMNIST                   & 0.87$\pm$0.00 & 0.69$\pm$0.00 & 0.87$\pm$0.02 & 0.99$\pm$0.02 & \good{0.80$\pm$0.06}  & \good{0.99$\pm$0.00}  & \bad{0.51$\pm$0.11} & \good{0.84$\pm$0.01} & \good{0.99$\pm$0.00}  & \bad{0.51$\pm$0.05}\\
            CIFAR10                  & 0.82$\pm$0.00 & 0.82$\pm$0.00 & 0.82$\pm$0.00 & 0.97$\pm$0.02 & \good{0.78$\pm$0.00}  & \good{0.97$\pm0.01$}  & \bad{0.65$\pm$0.01} & \good{0.80$\pm$0.01} & \good{0.90$\pm$0.01}  & \bad{0.69$\pm$0.01}\\\hline
            % \multirow{3}{*}{Dataset} & \multicolumn{2}{c|}{\advt}    & \multicolumn{8}{c|}{\radioactive}                                                                      \\
            %                          & \multicolumn{2}{c|}{Baseline} & \multicolumn{2}{c|}{Baseline}   & \multicolumn{3}{c|}{with \advt}                                     & \multicolumn{3}{c|}{with \advt Relaxed}\\
            %                          & \metricacc & \metricadv       & \metricacc & \metricrad         & \metricacc           & \metricrad            & \metricadv           & \metricacc            & \metricwm             & \metricadv\\\hline
            % MNIST                    & 0.99       & 0.95             & 0.98       & 0.284$\pm$0.001    & \good{0.94$\pm0.01$} & \bad{0.001$\pm$0.001} & \good{0.95$\pm$0.01} & \\
            % FMNIST                   & 0.87       & 0.69             & 0.88$\pm$0.01 & 0.191$\pm$0.002 & \good{0.87$\pm0.02$} & \bad{0.000$\pm$0.001} & \good{0.69$\pm$0.02} & \\
            % CIFAR10                  & 0.82       & 0.82             & 0.85        & 0.202$\pm$0.001   & \good{0.81$\pm0.01$} & \bad{0.003$\pm$0.002} & \good{0.81$\pm$0.01} & \\\hline
        \end{tabularx}
    % }
\end{table*}

\noindent\textbf{\radioactive with \advt}.
We observed a minor improvement in \metricacc and \metricadv.
However, \metricrad did not substantially improve (\Cref{tab:relaxed-advt-radioactive}).
The pair remains in conflict.

\setlength{\extrarowheight}{.2em}
\begin{table*}[ht]
    \centering
    \caption{Training on \traindata with \advt and without on \raddata.
    The change does not result in any meaningful improvement. Green underlined indicates no conflict; red indicates conflict.}
    \label{tab:relaxed-advt-radioactive}
    % \resizebox{2.\columnwidth}!{
    \scriptsize
    \begin{tabularx}\linewidth{|Y|Y|Y|Y|Y|Y|c|Y|Y|c|Y|}\hline
            \multirow{3}{*}{Dataset} & \multicolumn{2}{c|}{\advt}    & \multicolumn{8}{c|}{\radioactive}                                                                                                                                           \\
                                     & \multicolumn{2}{c|}{Baseline} & \multicolumn{2}{c|}{Baseline}   & \multicolumn{3}{c|}{+\advt}                                         & \multicolumn{3}{c|}{+\advt Relaxed}                             \\
                                     & \metricacc & \metricadv       & \metricacc & \metricrad         & \metricacc           & \metricrad            & \metricadv           & \metricacc           & \metricrad            & \metricadv           \\\hline
            MNIST                    & 0.99$\pm$0.00 & 0.95$\pm$0.00 & 0.98$\pm$0.00 & 0.284$\pm$0.001 & \good{0.94$\pm0.01$} & \bad{0.001$\pm$0.001} & \good{0.95$\pm$0.01} & \good{0.94$\pm0.02$} & \bad{0.002$\pm$0.001} & \good{0.94$\pm$0.03} \\
            FMNIST                   & 0.87$\pm$0.00 & 0.69$\pm$0.00 & 0.88$\pm$0.01 & 0.191$\pm$0.002 & \good{0.87$\pm0.02$} & \bad{0.000$\pm$0.001} & \good{0.69$\pm$0.02} & \good{0.87$\pm0.01$} & \bad{0.002$\pm$0.002} & \good{0.69$\pm$0.02} \\
            CIFAR10                  & 0.82$\pm$0.00 & 0.82$\pm$0.00 & 0.85$\pm$0.00 & 0.202$\pm$0.001 & \good{0.81$\pm0.01$} & \bad{0.003$\pm$0.002} & \good{0.81$\pm$0.01} & \good{0.82$\pm0.02$} & \bad{0.004$\pm$0.001} & \good{0.81$\pm$0.02} \\\hline
        \end{tabularx}
    % }
\end{table*}

\noindent\textbf{\wm with \dpsgd}. For MNIST and FMNIST, \fv achieves high \metricacc and \metricwm, comparable to the baseline (\Cref{tab:relaxed-dpsgd}).
For CIFAR10, \metricwm improves significantly but remains low enough to declare a conflict.

However, it begs the question whether these models are still $(\epsilon, \delta)$-private.
The idea behind \dpsgd is that it provides private \emph{training} by restricting the updates to model's \emph{weights}.
On one hand, using a regular SGD for \wm, and \dpsgd for \traindata breaks this assumption.
On the other, pre-training on public data (without privacy) and fine-tunning on a private \traindata has become the de facto way of training more accurate private models~\cite{tramer2021scatterdp,kurakin2022dpatscale}.

Additionally, \diffpr is often further relaxed to consider only \emph{computationally restricted adversaries}~\cite{mironov2009computational}, which provides the guarantee only for realistic datasets as opposed to any \traindata.
\triggerset could be considered irrelevant from the privacy standpoint.

\setlength{\extrarowheight}{.3em}
\begin{table}[ht]
    \centering
    \caption{Training on \traindata with \dpsgd and without on \triggerset. We recover performance close to the baseline. Green underlined indicates no conflict; red indicates conflict. }
    \label{tab:relaxed-dpsgd}
    % \resizebox{0.6\linewidth}!{
    \scriptsize
    \begin{tabularx}\columnwidth{|Y|Y|Y|Y|Y|Y|}\hline
            \multirow{3}{*}{Dataset} & \multicolumn{5}{c|}{\wm}                                                  \\
                                     & Baseline         & \multicolumn{2}{c|}{+\dpsgd}               & \multicolumn{2}{c|}{+\dpsgd Relaxed} \\
                                     &  \metricwm     & \metricacc           & \metricwm           & \metricacc           & \metricwm \\\hline
            MNIST                    &  0.97$\pm$0.01 & \good{0.97$\pm$0.00} & \bad{0.36$\pm$0.06} & \good{0.97$\pm0.01$} & \good{0.97$\pm$0.01}\\
            FMNIST                   &  0.99$\pm$0.02 & \good{0.86$\pm$0.00} & \bad{0.30$\pm$0.05} & \good{0.87$\pm0.01$} & \good{0.99$\pm$0.02}\\
            CIFAR10                  &  0.97$\pm$0.02 & \good{0.38$\pm$0.01} & \bad{0.12$\pm$0.01} & \good{0.39$\pm0.02$} & \bad{0.67$\pm$0.04}\\\hline
        \end{tabularx}
    % }
\end{table}

%% file: 8discussion.tex
%auto-ignore

\section{Discussion}
\label{sec:discussion}

\techreport{
We want to emphasize that our goal is not to criticize the techniques used throughout this work.
In our evaluation, they all perform ``as advertised'' for their intended purposes.
Rather, we highlight the conflicts that may arise in practice when two (or more) protection mechanisms that are effective by themselves, are deployed together on the same model.}

\subsection{Model Size \& Convergence}
\label{sec:model-size}
Insufficient model capacity and lack of convergence could be the source of the conflicts.
The conflict between \radioactive and \advt arises because \advt prevents watermarks from being embedded, and is thus independent of model size. In our experiments, all models reach low training loss, and expected accuracy.
Hence, larger models are unlikely to resolve the conflict between \advt and \wm.
Finally, for \diffpr, larger models deplete the privacy budget faster leading to lower accuracy.

\subsection{Other Mechanisms}
\label{sec:discussion-relaxations-schemes}

\dpsgd is the most popular mechanism for \diffpr training, but not the best performing one.
A recently proposed mechanism, ScatterDP~\cite{tramer2021scatterdp}, relies on training a classifier (logistic regression or small CNN) with \dpsgd on top of features in the frequency domain, obtained by transforming images with a ScatterNet~\cite{oyallon2015scatter}.
A small classifier does not have enough capacity to embed a watermark, and is more robust to perturbed inputs.
Using a bigger CNN removes the benefit of using ScatterDP.
Therefore, we deem that ScatterDP conflicts with ownership verification mechanisms because it does not admit joint deployment.

We do not evaluate pre-training on public data or any mechanisms that require it (e.g. PATE~\cite{papernot2018pate}).
Use of public data is not realistic in many industries, and has been primarily used for general purpose image and text models.
For instance, healthcare data typically cannot be disclosed due to the privacy regulation;
financial institutions have lengthy and restrictive compliance procedures.

\subsection{Limitations of Protection Mechanisms}
\label{sec:limitations-and-fps}

Most protection mechanisms evaluated in this work were previously shown to fall short when faced with a strong \adv.
\wm can be removed or prevented from being embedded~\cite{lukas2021wmsok}.
\advt does not generalise to higher $\gamma$ values~\cite{nie2022diffpure}.
More generally, attacks and defences against model evasion are defeated by novel approaches~\cite{carlini2017evadingadv,dixit2021poisoning}.
\diffpr requires careful, \textit{a priori} assumptions which often are not realistic~\cite{domingo2021dpmisuse}, and was recently shown to be vulnerable to side-channel timing attacks~\cite{jiankai2022dptiming}.

We also observed that \di results in false positives for models independently trained on a datset with the same distribution as \fv{'s} \traindata even if it \emph{does not} overlap with \fv{'s} \traindata.
Consequently, \di may result in innocent parties being falsely accused of stealing \fv.
We discuss this in more detail in~\Cref{app:fps}.
While \di avoids conflicts with other protection mechanisms we studied so far, we caution against using \di in domains where uniqueness of \fv's training data cannot be guaranteed.

\subsection{Stakeholders in the Training Loop}
\label{sec:discussion-stakeholders}

In a simple setting, a single party gathers the data, trains and deploys the model.
Hence, if \victim cares about data or model ownership they could decide to forgo \advt or \diffpr.

However, as ML services increasingly specialise, it is likely that different parties will be responsible for gathering data, providing the training platform, deploying the model, and using it.
In such a setting, even if the party deploying a model may not care about traceability with \radioactive, another involved party may.
Similarly, the training platform provider may want to embed a watermark to ensure that users of \fv conform to the terms of service, and not e.g. share it with others, or offer their own competing service using a knock-off of \fv.

We can consider a scenario where \victim concerned about model evasion, buys data from a party that uses \radioactive, and \fv is trained by some service that embeds a watermark.
\advt conflicts both with \radioactive and \wm.
Hence, data/platform provider needs to communicate up front that their offering is not compatible with certain training strategies, or resort to changes discussed in~\Cref{sec:discussion-relaxations-schemes}.

\subsection{Combinatorial Explosion of Simultaneous Comparisons}
\label{sec:discussion-combinatorial-explosion}

This work could be further extended to include triples or quadruples of protection mechanisms simultaneously.
Although, some of them could be considered toy cases, there are many combinations that reflect actual deployment considerations, e.g. \diffpr, \advt, \wm, while ensuring fairness.

However, increasing the size of the tuple leads to a \emph{combinatorial explosion} of the number of ways we can combine the protection mechanisms.
This number is likely to grow as new types of vulnerabilities are discovered, and does not account for multiple mechanisms within a single category.

%% file: 9related.tex
%auto-ignore

\section{Related Work}
\label{sec:related-work}

We summarise the prior work that studies interactions between properties in the context of ML security/privacy.
In~\Cref{app:all-conflicts}, we provide a broader overview of all unexplored pair-wise interactions that could result in conflicts.

It was shown that \diffpr can be used to certify robustness to model evasion~\cite{lecuyer2019dpandadv} by limiting the contribution of an individual pixel.
Prior work has extensively proved that using \diffpr degrades fairness of the models and can exacerbate bias present in the dataset as well as the performance on the downstream tasks~\cite{chang2021privacyalgofairness,cheng2021synthfairness,pearce2022privandfair}.
Membership inference attacks (MIAs) were used to evaluate the privacy guarantee of \dpsgd~\cite{milad2021advinst}, and it was argued that \diffpr should provide resistance to them~\cite{humpfries2020miadp}.

It was suggested that poisoning attacks can be used to make models vulnerable to other threats.
One can inject samples into the training set to make MIAs easier~\cite{tramer2022truthserum}.
Also, there is a connection between adversarial robustness and susceptibility to poisoning~\cite{pang2020eviltwins}.
Furthermore, adversarial robustness can make models more vulnerable to MIAs~\cite{song2019privadvtr}.

Nevertheless, it was shown that representations learnt with \advt can make models more interpretable~\cite{tsipras2019robustness}.
Lastly, LIME~\cite{ribeiro2016lime}, a popular explainability method, was used to compare the similarity of models~\cite{jia2022lime}.
However, it was shown that post-hoc explainability methods can be used to speed up model evasion, model extraction, and membership inference~\cite{quan2022adhocrisk}.

%% file: 10conclusion.tex
%auto-ignore
\section{Conclusion}
\label{sec:conclusion}

In this work, we pose the problem of \emph{conflicting interactions} between different ML protection mechanisms.
We provide a framework for evaluating simultaneous deployment of multiple mechanisms.
We use it explore the interaction between three ownership verification mechanisms (\wm, \di, \radioactive) with the most popular methods for preventing model evasion (\advt), and differentially private training (\dpsgd).
We show there exists a theoretical and empirical conflict that limits the effectiveness of multiple mechanisms (\Cref{app:summary}).

Moving forward, researchers working on ML protection mechanisms should extend their evaluation benchmarks to include conflicts with protection against other common concerns.
This in turn, would allow practitioners to choose the most appropriate mechanisms for their deployment scenario and threat model.
We emphasize that this is not merely an ``engineering problem'' that can be ignored during the research phase but a key consideration for any technique and its deployability prospects.

Certain considerations are relevant only to particular applications, e.g. fairness is at odds with privacy but it might not be important when used in a closed loop system.
Similarly, adversarial training may hurt data-based watermarking and fingerprinting mechanisms but is outside of the threat model of systems that do not have a user facing interface.

Many pairs still require explicit analysis that could unravel surprising limitations (\Cref{app:all-conflicts}).
We encourage the community to build upon this work, and extend it to additional pairs, and bigger tuples.

%% file: 11acknowledgements.tex
%auto-ignore
\section*{Acknowledgements}
\label{sec:acknowledgements}

We would like to thank Pui Yan Chloe Sham, Dmytro Shynkevych and Asim Waheed for contributing to the implementation during the early stages of this project. We thank Jason Martin whose remarks set us on the path of exploring conflicting interactions among ML protection mechanisms.
This work was supported in part by Intel (in the context of the Private-AI Institute).

%% file: 12appendix.tex
\appendix

\section{Impact of the Hyperparameters}
\label{app:hyperparams}

\begin{figure}[!h]
    \includegraphics[width=\columnwidth]{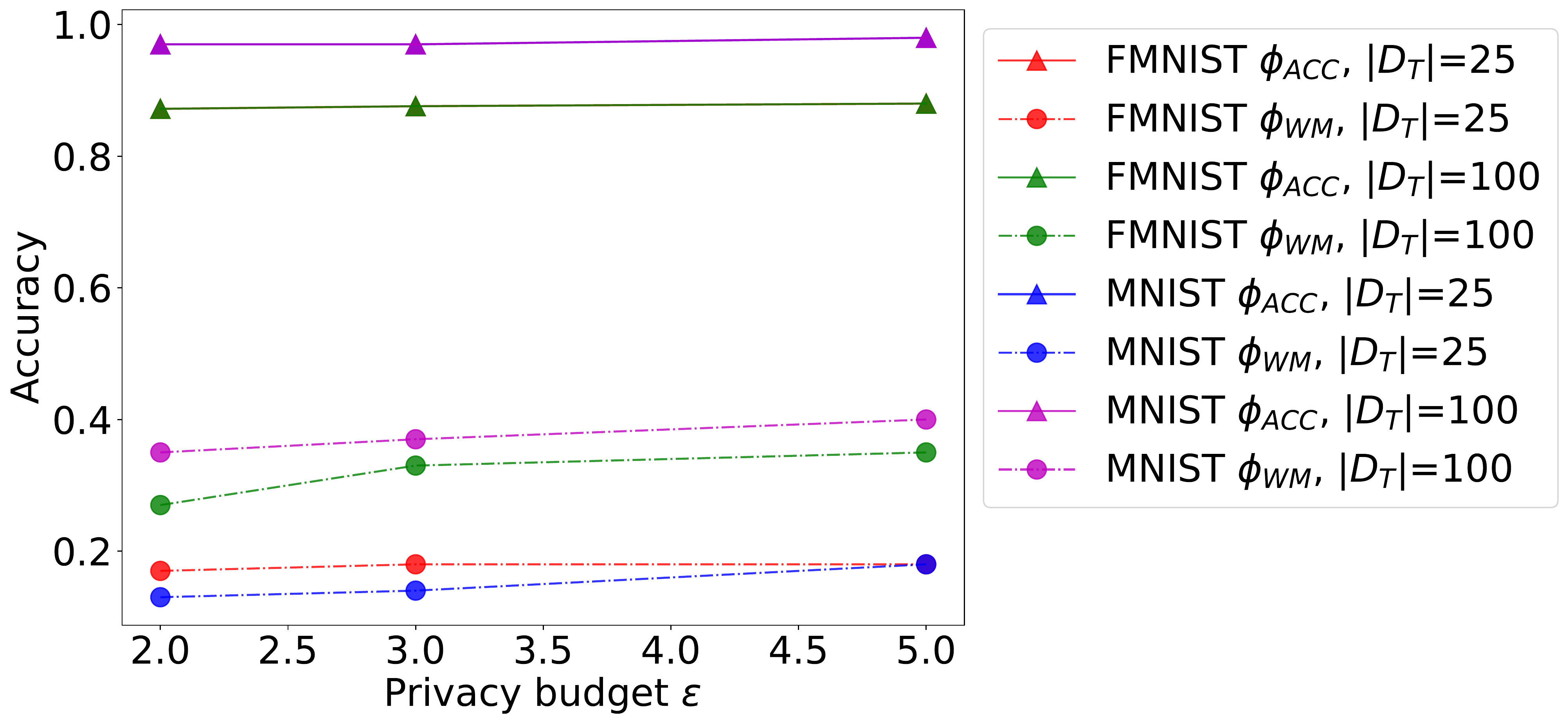}
    \caption{\textbf{\wm with \dpsgd}. Changing $\epsilon$ has small impact on \metricwm. However, we observe that bigger $|\triggerset|$ leads to higher \metricwm. We omit the error bars for visual clarity.}
    \label{fig:dp_params}
\end{figure}

\begin{table*}[]
    \centering
    \caption{The most common concerns in security and privacy of machine learning. For brevity, the label corresponds either to the attack or the protection mechanism.}
    \label{tab:all-properties}
    % \resizebox{2.\columnwidth}!{
    \resizebox{\linewidth}!{
    \begin{tabular}{| c | c | c | c | c | c | c | c | c | c | c | c | }\hline
    \multirow{2}{*}{Concern}              & Model                     & Differential                                       & Membership                                      & Oblivious                   & Model/Gradient               & Model                                           & Model                              & Model                           & Data                              & \multirow{2}{*}{Explainability}                     & \multirow{2}{*}{Fairness}\\
                                          & Evasion                   & Privacy                                            & Inference                                       & Training                    & Inversion                    & Poisoning                                       & Watermarking                       & Fingerprinting                  & Watermarking                      &                                                     &                          \\\hline
    \multirow{2}{*}{Model Evasion}        & \cecg                     &  \cecp                                             & \cecp                                           &                             &                              & \cecp                                           & \cecb                              & \cecb                           & \cecb                             & \cecp\cite{tsipras2019robustness}                   &                          \\
                                          & \multirow{-2}{*}{\cecg X} &  \multirow{-2}{*}{\cecp\cite{lecuyer2019dpandadv}} & \multirow{-2}{*}{\cecp\cite{song2019privadvtr}} & \multirow{-2}{*}{\textbf{?}}& \multirow{-2}{*}{\textbf{?}} & \multirow{-2}{*}{\cecp\cite{pang2020eviltwins}} & \multirow{-2}{*}{\cecb\ours}       & \multirow{-2}{*}{\cecb\ours}    & \multirow{-2}{*}{\cecb\ours}      & \cecp\cite{quan2022adhocrisk}                       & \multirow{-2}{*}{\textbf{?}}           \\
    \multirow{2}{*}{Differential Privacy} &                           &  \cecg                                             & \cecp\cite{humpfries2020miadp}                  &                             &                              &                                                 & \cecb                              & \cecb                           & \cecb                             &                                                     & \cecp\cite{chang2021privacyalgofairness}\\
                                          &                           &  \multirow{-2}{*}{\cecg X}                         & \cecp\cite{milad2021advinst}                    & \multirow{-2}{*}{\textbf{?}}& \multirow{-2}{*}{\textbf{?}} & \multirow{-2}{*}{\textbf{?}}                    & \multirow{-2}{*}{\cecb\ours}       & \multirow{-2}{*}{\cecb\ours}    & \multirow{-2}{*}{\cecb\ours}      & \multirow{-2}{*}{\textbf{?}}                        & \cecp\cite{cheng2021synthfairness,pearce2022privandfair}\\
    Membership Inference                  &                           &                                                    & \cecg X                                         & \textbf{?}                  & \textbf{?}                   & \cecp\cite{tramer2022truthserum}                & \textbf{?}                         &  \textbf{?}                     & \textbf{?}                        & \textbf{?}                                          & \textbf{?}               \\
    Oblivious Training                    &                           &                                                    &                                                 & \cecg X                     & \textbf{?}                   & \textbf{?}                                      & \textbf{?}                         &  \textbf{?}                     & \textbf{?}                        & \textbf{?}                                          & \textbf{?}               \\
    Model/Gradient Inversion              &                           &                                                    &                                                 &                             &  \cecg X                     & \textbf{?}                                      & \textbf{?}                         &  \textbf{?}                     & \textbf{?}                        & \textbf{?}                                          & \textbf{?}               \\
    Model Poisoning                       &                           &                                                    &                                                 &                             &                              & \cecg X                                         & \textbf{?}                         &  \textbf{?}                     & \textbf{?}                        & \textbf{?}                                          & \textbf{?}               \\
    Model Watermarking                    &                           &                                                    &                                                 &                             &                              &                                                 & \cecg X                            &  \textbf{?}                     & \textbf{?}                        & \textbf{?}                                          & \textbf{?}               \\
    \multirow{2}{*}{Model Fingerprinting} &                           &                                                    &                                                 &                             &                              &                                                 &                                    &   \cecg                         &                                   & \cecp\cite{jia2022lime}                             &                          \\
                                          &                           &                                                    &                                                 &                             &                              &                                                 &                                    &  \multirow{-2}{*}{\cecg X}      & \multirow{-2}{*}{\textbf{?}}      & \cecp\cite{quan2022adhocrisk}                       & \multirow{-2}{*}{\textbf{?}}\\
    Data Watermarking                     &                           &                                                    &                                                 &                             &                              &                                                 &                                    &                                 &  \cecg X                          & \textbf{?}                                          & \textbf{?}               \\
    Explainability                        &                           &                                                    &                                                 &                             &                              &                                                 &                                    &                                 &                                   &          \cecg X                                    & \textbf{?}               \\
    Fairness                              &                           &                                                    &                                                 &                             &                              &                                                 &                                    &                                 &                                   &                                                     &       \cecg X            \\
    \hline
    \end{tabular}}
\end{table*}

\begin{figure*}[!ht]
    \resizebox{1.\linewidth}!{
    \begin{tabular}{cc}
    \includegraphics[width=0.33\columnwidth]{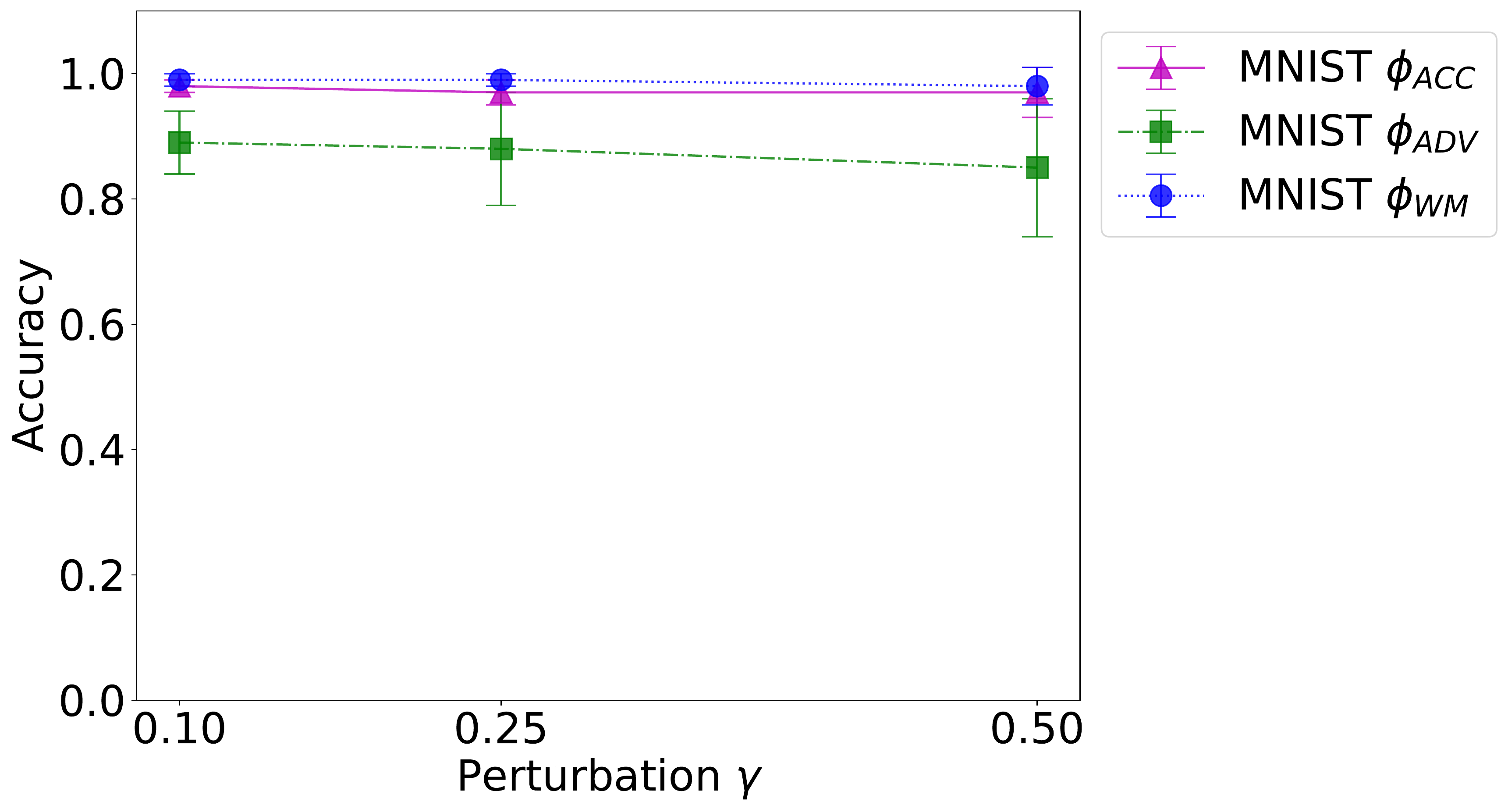}
    \includegraphics[width=0.33\columnwidth]{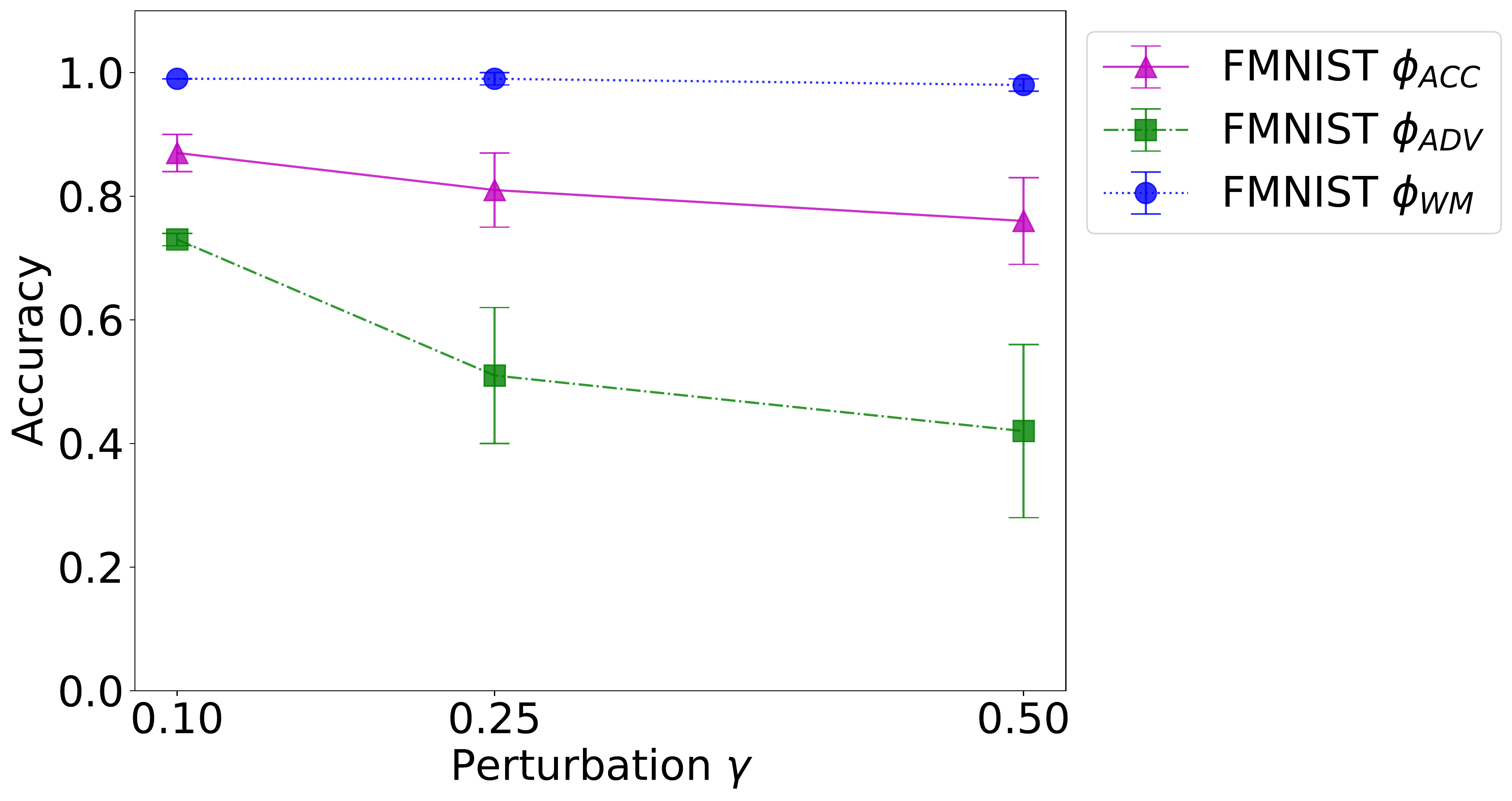}
    \includegraphics[width=0.33\columnwidth]{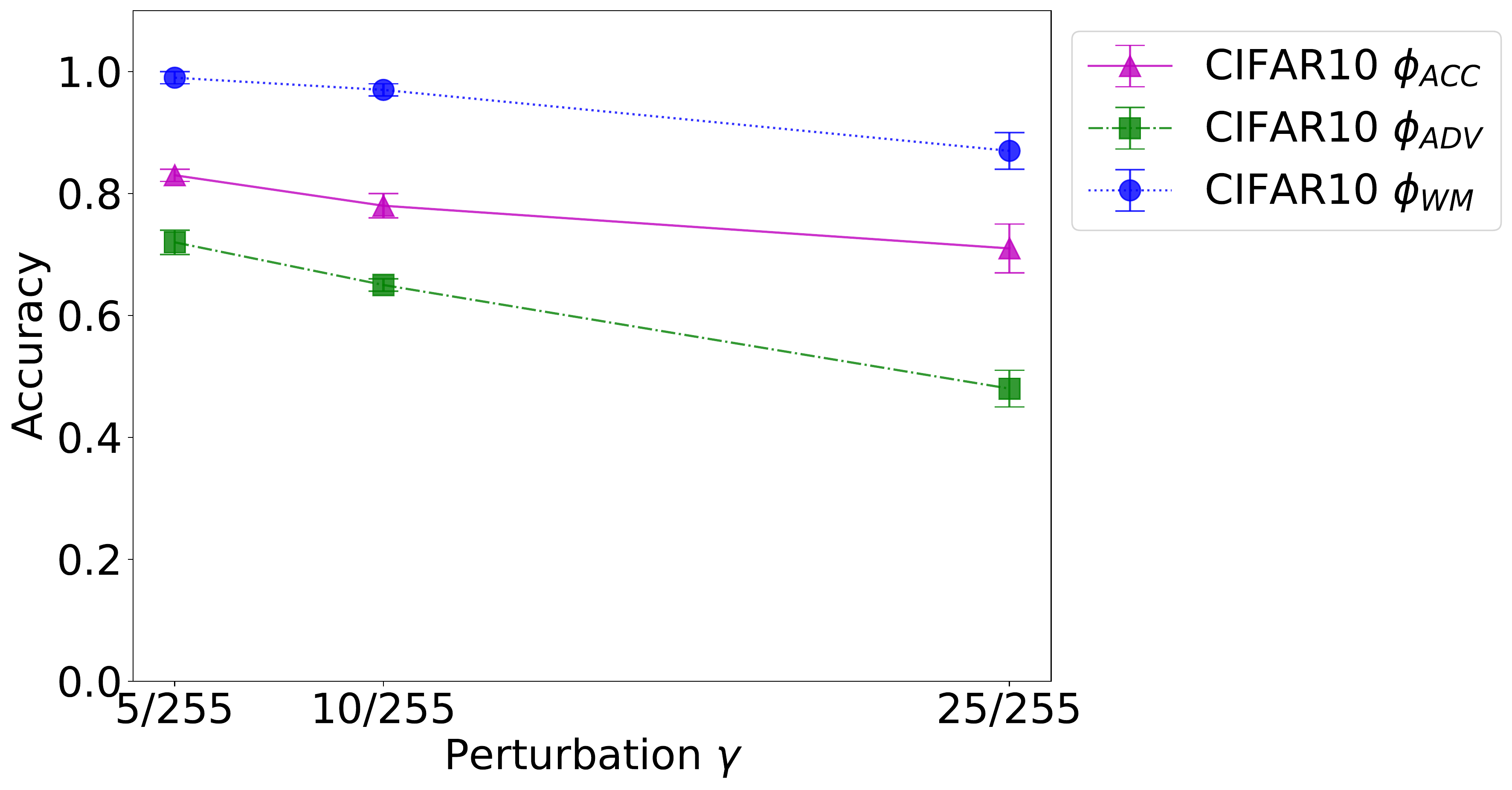}
    \end{tabular}}
    \caption{\textbf{\wm with \advt}. Changing $\gamma$ has marginal impact on \metricwm.
    For lower $\gamma$, FMNIST is not in conflict, while CIFAR10 is.
    \metricadv measured with the corresponding}
    \label{fig:eps}
\end{figure*}

\noindent\textbf{\wm with \advt}.
We test different values of $\gamma$ for \advt to check if there exists a sweet spot that gives acceptable performance for both.
We observe that increasing $\gamma$ has disproportionate impact on \metricacc and \metricadv (\Cref{fig:eps}).
On the other hand, for the lower value of $\gamma=0.1$, FMNIST is no longer in conflict.
Nevertheless, \victim must accept much weaker protection as \advt does not provide any meaningful protection against adversarial examples crafted with $\gamma$ higher than the one used during training.
CIFAR10 remains in conflict.

\begin{figure*}[!ht]
    \resizebox{1.\linewidth}!{
    \begin{tabular}{cc}
    \includegraphics[width=0.33\columnwidth]{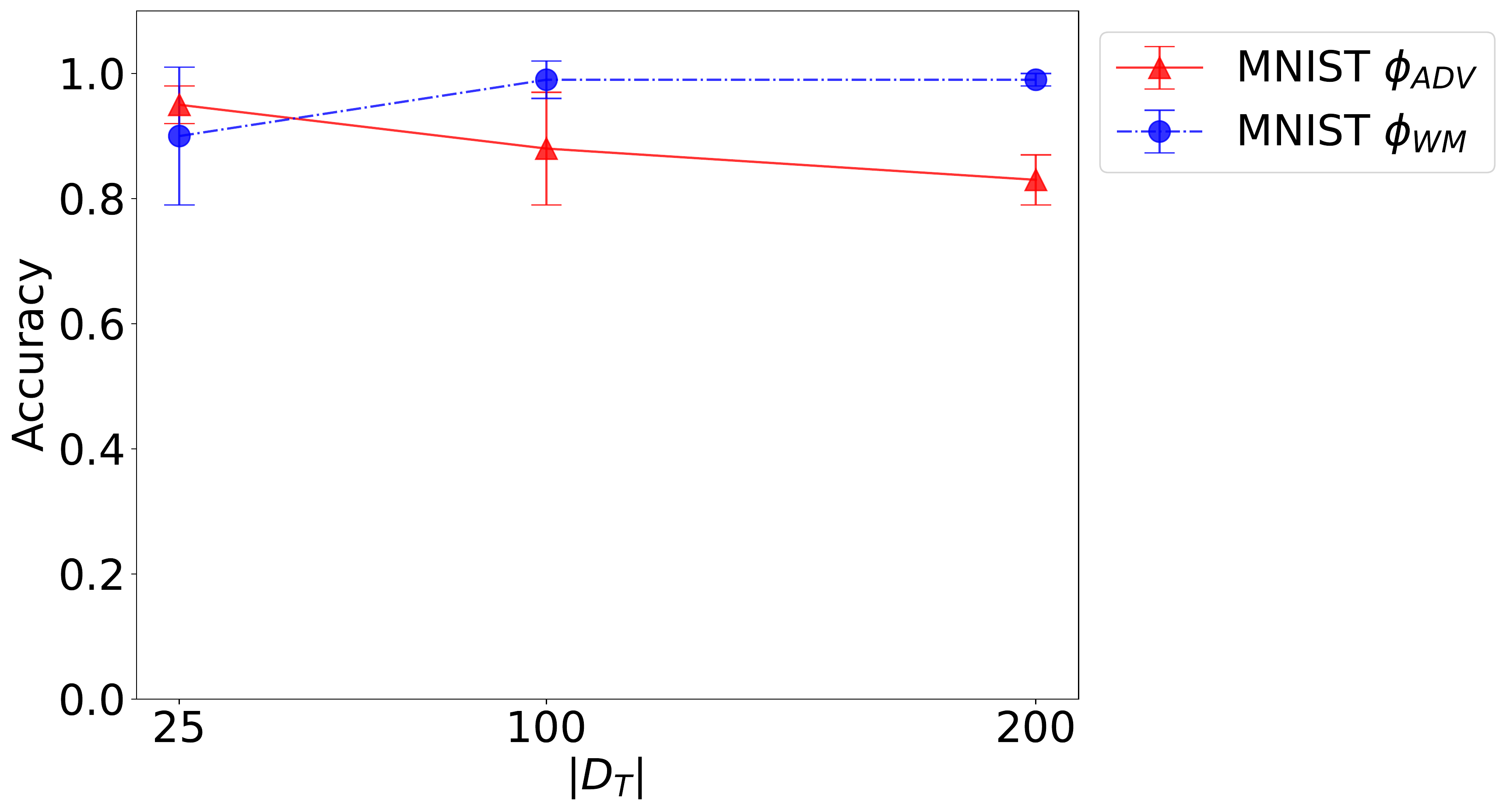}
    \includegraphics[width=0.33\columnwidth]{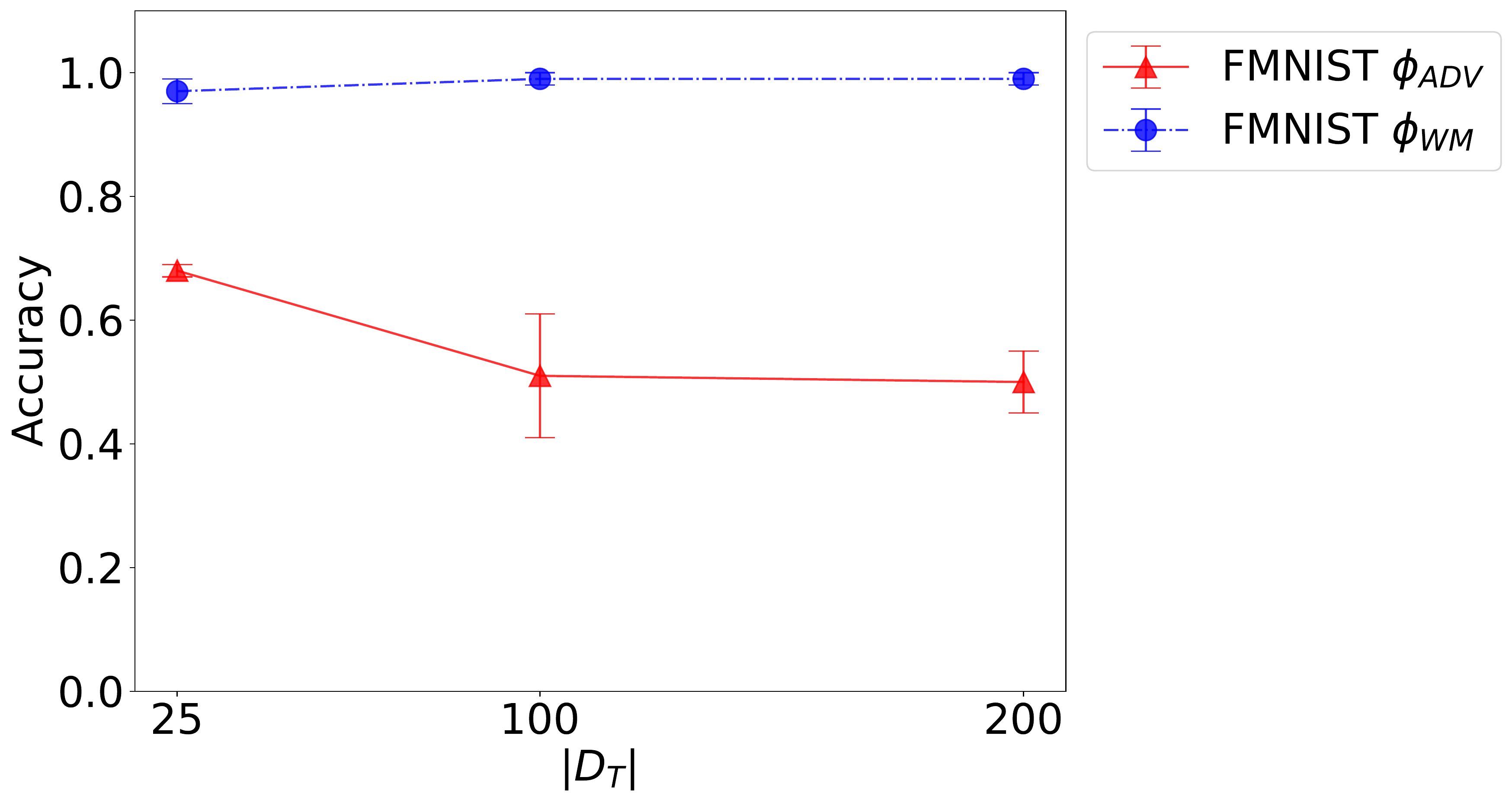}
    \includegraphics[width=0.33\columnwidth]{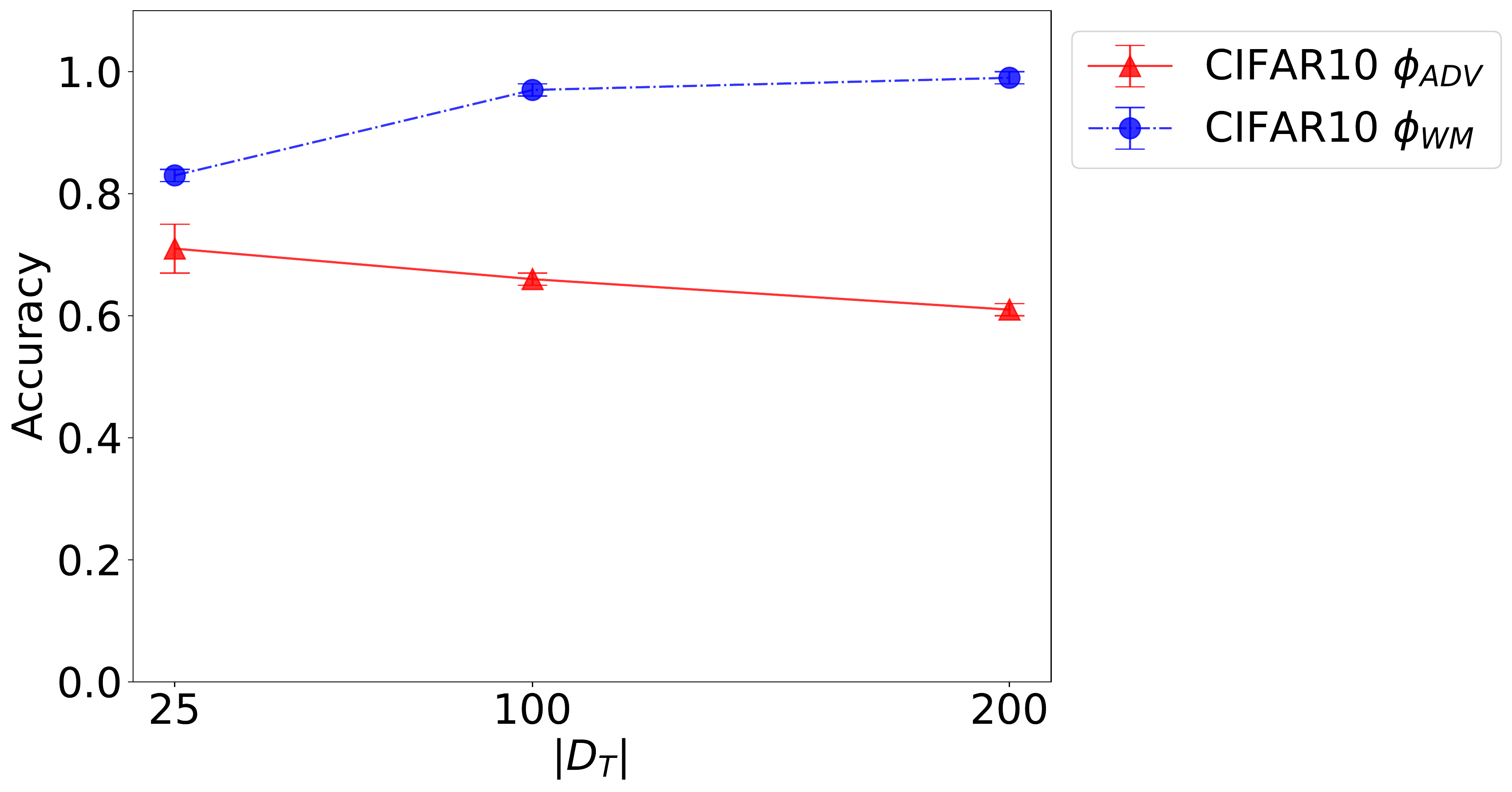}
    \end{tabular}}
    \caption{\textbf{\wm with \advt}.
    Decreasing $|\triggerset|$ improves \metricadv and worsens \metricwm, while increasing it has the opposite effect.
    For $|\triggerset|=25$, MNIST has no conflict with \advt.
    FMNIST and CIFAR10 remain in conflict.}
    \label{fig:dt}
\end{figure*}

On the other hand, while decreasing \triggerset improves \metricadv, it worsens \metricwm (\Cref{fig:dt}).
Smaller \triggerset and lower \metricwm together reduce the confidence of the watermark (Equation 2), e.g. for CIFAR10 changing \triggerset from $100$ to $25$, leads to a significant decrease ($(1 - 10^{-92}) \rightarrow (1 - 10^{-18})$).
FMNIST and CIFAR10 remain in conflict.

\noindent\textbf{\wm with \dpsgd}.
We test different $\epsilon$ and $|\triggerset|$ values to check if we can improve on the baseline result~\Cref{fig:dp_params}
Although increasing $\epsilon$ improves \metricwm, we observe that for all three values of $\epsilon$, $\metricwm < 0.5$.
Varying $|\triggerset|$ has a more noticeable impact on \metricwm.
Lowering $|\triggerset|=25$, leads to \metricwm lower than for $|\triggerset|=100$.
Nevertheless, all three datasets remain in conflict.

\noindent\textbf{\radioactive with \advt}.
We test larger $|\raddata|$ to check if it improves \metricrad.
We evaluate $\raddata$ that constitutes $20\%$ and $35\%$ of \traindata.

When used without \advt, larger \raddata leads to higher \metricrad at the cost of \metricacc.
With \advt, \metricrad remains close to zero, and \metricacc drops.
Hence, this makes the conflict even worse.

\section{False Positives in \di}
\label{app:fps}

In our experiments with \di, we identified a potential false positive issue.
The mechanism flags a model that was trained on the same distribution of data as a stolen model even if it does not overlap with the training data.
Therefore, \di can be used to frame innocent parties that offer similar models to \fv, and that were trained on a dataset distinct from \fv{'s} training data.

We conducted the following experiment: 1) divide CIFAR10 into two random, equally sized chunks A and B 2) chunk A is \victim{'s} new training set, while chunk B is the training set of an independent party.
We observe that \di falsely classifies the model trained on the chunk B as stolen with high confidence (\Cref{tab:di-fp}).

\begin{table}[]
    \centering
    \caption{Simulated experiment with two non-overlapping chunks of CIFAR10 training data. Independent model (chunk B) triggers a false positive during ownership verification.}
    \label{tab:di-fp}
    \begin{tabularx}\linewidth{|Y|Y|}\hline
        Training Data      & \metricdi \\ \hline
        Chunk A (\victim)  & $\good{10^{-23}}$ \\
        Test (independent) & $\good{0.1}$ \\
        Chunk B (framed)   & $\bad{10^{-12}}$ \\ \hline
    \end{tabularx}
\end{table}

Even though in the threat model of \di, it was argued that \victim has a \emph{private dataset}, we argue it is difficult to guarantee that an independent dataset will have sufficiently different distribution from the private dataset.
Furthermore, two parties may have bought their data from the same vendor that generates a unique synthetic dataset for each of their clients.
While it does not break this mechanism entirely, we caution against using \di in domains where uniqueness of data cannot be guaranteed.

\section{Possible Pair-wise Conflicts}
\label{app:all-conflicts}

In~\Cref{tab:all-properties} we give an overview of prior work that studies interactions between different properties in the context of ML security/privacy.
Note that these can be divided into two categories: 1) attacks or defences that make \fv more vulnerable to a threat; 2) properties or defences that may be incompatible during simultaneous deployment.

\section{Summary of Conflicts}
\label{app:summary}

In \Cref{tab:summary} we provide an overview of all conflicts identified in this work.
Out of six pairs, there are three conflicting interaction: \wm with \diffpr, \wm with \advt, and \radioactive with \advt.

\begin{table}[]
    \centering
    \caption{Summary of the conflicts identified in this work. Red \metric indicates that it is the cause of the conflict; we colour a cell red if any conflict occurs. Green (underlined) indicates no conflict.}
    \label{tab:summary}
    \scriptsize
    \begin{tabularx}\linewidth{|Y|Y|c|c|}\hline
        Protection                    & \multirow{2}{*}{Dataset} & \multicolumn{2}{c|}{Protection Mechanism} \\\cline{3-4}
        Mechanism                     &                          & \diffpr                                   & \advt\\\hline
        \multirow{3}{*}{\wm}          & MNIST                    & \cecp\good{\metricacc} \bad{\metricwm}    & \cecgr\good{\metricacc} \good{\metricwm} \good{\metricadv}\\
                                      & FMNIST                   & \cecp\good{\metricacc} \bad{\metricwm}    & \cecp\good{\metricacc} \good{\metricwm} \bad{\metricadv}\\
                                      & CIFAR10                  & \cecp\good{\metricacc} \bad{\metricwm}    & \cecp\good{\metricacc} \good{\metricwm} \bad{\metricadv}\\\hline
        \multirow{3}{*}{\radioactive} & MNIST                    & \cecgr\good{\metricacc} \good{\metricrad} & \cecp\good{\metricacc} \bad{\metricrad} \good{\metricadv}\\
                                      & FMNIST                   & \cecgr\good{\metricacc} \good{\metricrad} & \cecp\good{\metricacc} \bad{\metricrad} \good{\metricadv}\\
                                      & CIFAR10                  & \cecgr\good{\metricacc} \good{\metricrad} & \cecp\good{\metricacc} \bad{\metricrad} \good{\metricadv}\\\hline
        \multirow{3}{*}{\di}          & MNIST                    & \cecgr\good{\metricacc} \good{\metricdi}  & \cecgr\good{\metricacc} \good{\metricdi} \good{\metricadv}\\
                                      & FMNIST                   & \cecgr\good{\metricacc} \good{\metricdi}  & \cecgr\good{\metricacc} \good{\metricdi} \good{\metricadv}\\
                                      & CIFAR10                  & \cecgr\good{\metricacc} \good{\metricdi}  & \cecgr\good{\metricacc} \good{\metricdi} \good{\metricadv}\\\hline
    \end{tabularx}
\end{table}